\documentclass[journal]{IEEEtran}
\usepackage{amssymb}
\usepackage{booktabs}
\usepackage{amsmath}
\usepackage{graphicx}
\usepackage{diagbox} 
\usepackage{stfloats}
\usepackage{array} 
\usepackage{color}
\usepackage{subfigure}
\usepackage{threeparttable}

\usepackage{url}
\usepackage{float}

\usepackage[numbers,sort&compress]{natbib}
\usepackage{makecell,rotating}
\usepackage{multirow}

\usepackage{lineno}

\begin{document}

\newpage
\mbox{This work has been submitted to the IEEE for possible publication. Copyright may be transferred without notice, after}
\mbox{which this version may no longer be accessible.}
\newpage

\title{Smoothness Sensor: Adaptive Smoothness-Transition Graph Convolutions for Attributed Graph Clustering
}

\author{Chaojie~Ji,
        Hongwei~Chen,
        Ruxin~Wang,
        Yunpeng~Cai,
        and~Hongyan~Wu 
\thanks{Manuscript received XXXX XX, XXXX; revised XXXX XX, XXXX. (\emph{Corresponding author: Hongyan Wu.})}
\thanks{Chaojie Ji, Ruxin Wang, Yunpeng Cai and Hongyan Wu are with the Joint Engineering Research Center for Health Big Data Intelligent Analysis Technology, Shenzhen Institutes of Advanced Technology, Chinese Academy of Sciences, Shenzhen, China (e-mail: cj.ji@siat.ac.cn; rx.wang@siat.ac.cn; yp.cai@siat.ac.cn; hy.wu@siat.ac.cn).}
\thanks{Hongwei Chen is with the Shenzhen Institutes of Advanced Technology, Chinese Academy of Sciences, Shenzhen, China, and University of Chinese Academy of Sciences, Beijing, China (e-mail: hw.chen@siat.ac.cn).}
\thanks{Chaojie Ji and Hongwei Chen make equal contribution.}}

\markboth{IEEE, VOL. XX, NO. XX, XXXX XXXX}
{Shell \MakeLowercase{\textit{et al.}}: Bare Demo of IEEEtran.cls for IEEE Journals}

\maketitle

\begin{abstract}
Clustering techniques attempt to group objects with similar properties into a cluster. Clustering the nodes of an attributed graph, in which each node is associated with a set of feature attributes, has attracted significant attention. Graph convolutional networks (GCNs) represent an effective approach for integrating the two complementary factors of node attributes and structural information for attributed graph clustering.
However, oversmoothing of GCNs produces indistinguishable representations of nodes, such that the nodes in a graph tend to be grouped into fewer clusters, and poses a challenge due to the resulting performance drop.
In this study, we propose a smoothness sensor for attributed graph clustering based on adaptive smoothness-transition graph convolutions, which senses the smoothness of a graph and adaptively terminates the current convolution once the smoothness is saturated to prevent oversmoothing.
Furthermore, as an alternative to graph-level smoothness, a novel fine-gained node-wise level assessment of smoothness is proposed, in which
smoothness is computed in accordance with the neighborhood conditions of a given node at a certain order of graph convolution.
In addition, a self-supervision criterion is designed considering both the tightness within clusters and the separation between clusters to guide the whole neural network training process.
Experiments show that the proposed methods significantly outperform 12 other state-of-the-art baselines in terms of three different metrics across four benchmark datasets. In addition, an extensive study reveals the reasons for their effectiveness and efficiency.
\end{abstract}


\section{Introduction}
\IEEEPARstart{C}{lustering} techniques attempt to group objects with similar properties into a cluster \cite{hartigan1979algorithm}, \cite{ng2002spectral}.Various methods have been proposed to solve real-world problems via text \cite{aggarwal2012survey} and image \cite{yang2010image} clustering.
Recently, with the emergence of graph-structured data such as social networks and biological networks \cite{lancichinetti2011finding, zachary1977information, ji2020graph}, the partitioning of the nodes of an attributed graph, in which each node is associated with a set of feature attributes, has attracted significant attention \cite{8123526}, \cite{8567993}. For example, potential criminal organizations can be identified based on frequent contacts among known criminals \cite{ferrara2014detecting}.
In a graph, attributes represent the feature values of a vertex itself, while the structural information indicates the underlying similarity among graph nodes, including not only the relationships within a one-hop distance but also more complex relationships at higher-order distances \cite{8766847}, \cite{8618599}. The question of how to effectively integrate these two complementary factors of attributes and structural information for the task of clustering an attributed graph has attracted the interest of researchers.

Classical data clustering algorithms, such as k-means, construct a similarity matrix of the node features and then perform clustering on this matrix \cite{von2007tutorial}. Network-structure-based approaches, e.g., \cite{newman2006finding}, use Laplacian eigenmaps to group nodes with a higher-than-average density of edges, emphasizing the importance of structural information. \cite{wang2016structural} exploits first-order and second-order proximity to jointly preserve the global and local structures of a network.
Although later researches attempt to integrate both node feature and network structure \cite{cai2008non}, \cite{gu2009co}, these methods less explore deep representation learning.

Recently, deep-learning-related methods have been exploited to learn graph representations based on both node content and network structure information \cite{wu2020comprehensive}. Classical clustering techniques, e.g., k-means and spectral clustering \cite{wang2017mgae}, can be stacked on the low-dimensional representations learned by deep learning networks.
These deep-learning-based methods can be roughly categorized into two classes: autoencoders and graph convolutional networks (GCNs).
Tian et al. first proposed a method for learning a nonlinear embedding of the original graph using stacked autoencoders \cite{tian2014learning}.
GCNs are designed to naturally incorporate information on the nodes themselves and the relationships among nodes \cite{kipf2016semi}. Variational graph autoencoders (VGAEs) and graph autoencoders (GAEs) rely on a graph convolutional network encoder and a simple inner product decoder for unsupervised learning.

Most of the existing methods rely on the application of fixed shallow (low-order) graph convolutions. To capture deep structural information, a structural deep clustering network (SDCN) \cite{sdcn2020} uses a delivery operator to transfer learned representations from an autodecoder to GCN layers to combine both low-order and high-order information.
Marginalized graph autoencoders (MGAEs) were developed in an attempt to extend autoencoders to deep convolutions to learn more effective representations \cite{wang2017mgae}. However, the number of layers in an MGAE is still limited to 3. Adaptive graph convolution (AGC) has been developed by introducing a novel low-pass graph filter and extending it to a $k$-order graph convolution \cite{zhang2019attributed}.

Smoothness is an indicator for assessing the degree of similarity of feature representations among nearby nodes in a graph. With higher orders of graph convolution, smoother filtered graph signals are obtained. Graph convolution with an excessively large order $k$ results in oversmooth node representations. Oversmoothing has been identified as a major cause of performance degradation in deep graph convolutional networks and the downstream tasks thereafter \cite{li2018deeper}.

\begin{figure*}[t]
 \centering
 \includegraphics[scale=0.5]{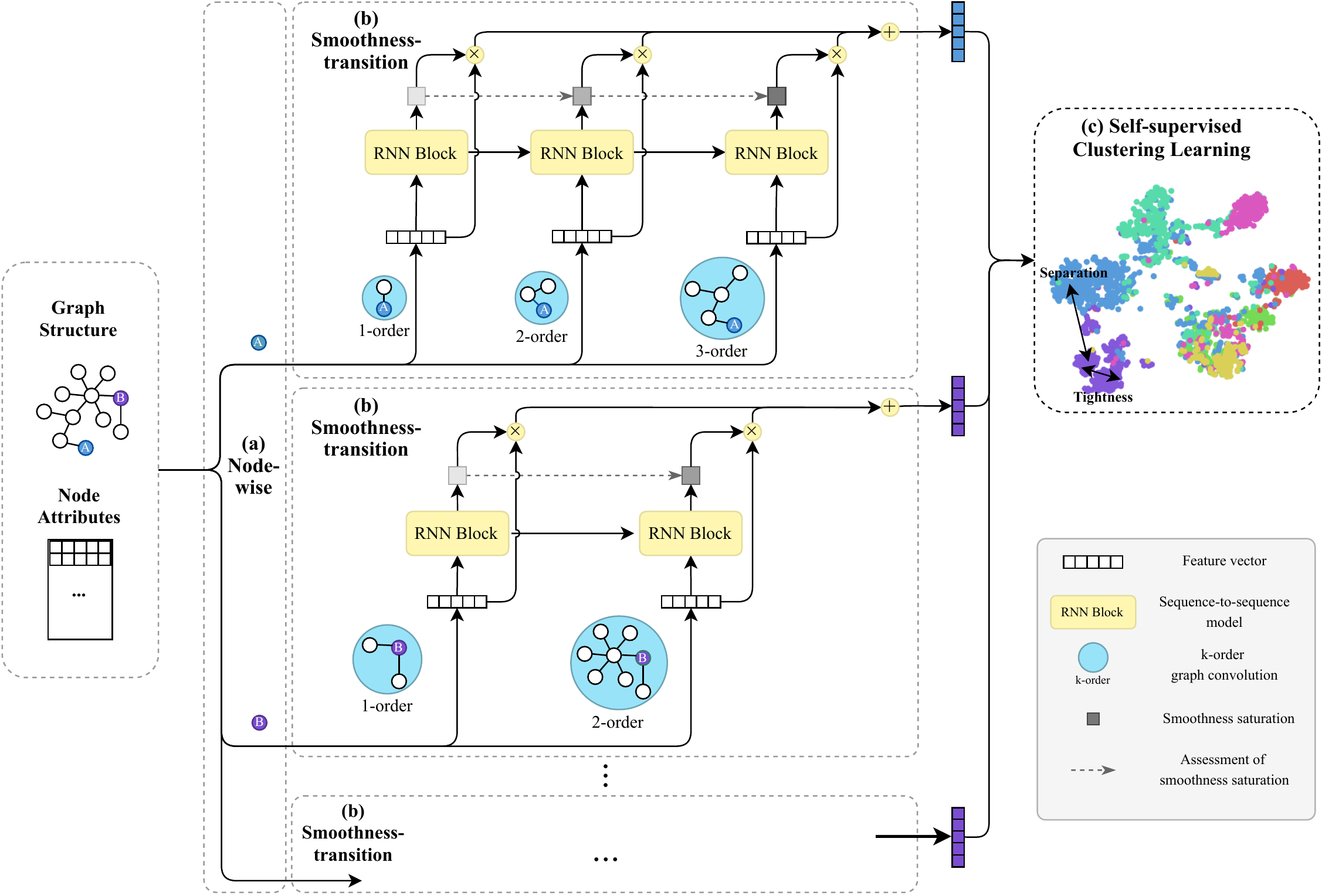}
 \caption{The framework of the proposed smoothness sensor of NAS-GC. (a) Perform convolution based on the node level and detect the fine-grained  saturation of smoothness; (b) the core smoothness-transition component operates to quantify the saturation of smoothness in the form of standard sequence-to-sequence prediction, considering a gradual smoothness process; (c) Once clustering partitioning is achieved by applying the k-means algorithm to the learned representations, the proposed self-supervised clustering strategy is employed to each node pair in accordance with both the tightness within clusters and the separation between clusters.}
 \label{overall_framework}
\end{figure*}

In addition, all of the existing methods roughly specify a certain order $k$ of graph convolution for all nodes in the whole graph. However, the node density can vary greatly in attributed graphs. A relatively isolated node with few neighbors could require a larger $k$ (stronger smoothing) to introduce more distant nodes to obtain more information, while a node with more neighbors usually can efficiently gather information with a smaller $k$ (weaker smoothing). Thus, fixed $k$-order graph convolution at the graph level is clearly a coarse solution.

In this study, to address these two problems, we propose an adaptive smoothness-transition graph convolution method for attributed graph clustering as Fig. \ref{overall_framework}, which operates like a \textbf{smoothness sensor}. In particular, the smoothness is sensed at the fine-grained node level instead of at the level of the entire graph. Finally, a self-supervision strategy is designed to enable the training of our proposed model with respect to various graph structures.

\textbf{Smoothness-transition adaptivity}. To overcome the oversmoothing of GCNs, it is necessary to adaptively customize the order $k$ of graph convolution. We explore how $k$-order filtered graph signals are transited from $(k-1)$-order filtered signals, and we define an assessment of smoothness saturation related to iterative graph convolution operations as a standard sequence-to-sequence prediction problem, in which the saturation of smoothness is taken as an indicator of whether the convolution process should be terminated as shown in Fig. \ref{overall_framework} (b).

\textbf{Fine-grained smoothness}. To further detect the saturation of smoothness at a fine-grained level that can adaptively consider the surrounding environment of each node, we first introduce a preliminary model --- Adaptive Smoothness-transition Graph Convolution (AS-GC) --- for detecting graph-wise smoothness and then evolve it into a mature version based on node-wise smoothness, that is, Node-wise Adaptive Smoothness-transition Graph Convolution (NAS-GC) as illustrated in Fig. \ref{overall_framework} (a).

\textbf{Self-supervised clustering criterion}. Clustering is a typical unsupervised learning problem. To adapt traditional GCNs, which requires semi-supervision, to clustering scenarios, we propose a complete and compact clustering criterion that can directly consider both the tightness within clusters and the separation between clusters to guide the self-supervised learning process of NAS-GC (AS-GC), which is integrated as a component of self-supervised clustering learning shown in Fig. \ref{overall_framework} (c).

We conduct a series of experiments to investigate the proposed approaches. The experiments prove that the proposed methods significantly outperform 12 other state-of-the-art baselines in terms of three popular metrics across four benchmark datasets. Our extensive study shows the effectiveness and efficiency of our proposed methods and further reveals their implicit mechanisms.

The remaining of this paper is organized as follows: Section 2 reviews the related work of attributed graph clustering in terms of machine learning and deep learning. We elaborate on the details of the procedure of evolution and details of the proposed AS-GC and NAS-GC. In section 4, experimental results are given for evaluating the proposed models with respect to effectiveness and efficiency and an extensive study is conducted to reveal the intrinsic mechanism of our methods.

\section{Related Work}
Our work focuses on attributed graphs, in which every node is associated with a set of feature attributes and the nodes are connected to each other \cite{cai2018comprehensive}, \cite{yang2009combining}. Then, these nodes are clustered in accordance with both their feature attributes and the structural information of the graph \cite{8118287}. Related methods can be mainly categorized into two classes --- machine-learning- and deep-learning-based representations.

\subsection{Graph Clustering through Machine Learning}
A series of classical methods based on node features have been proposed.
Although the graph structure, in some cases, is not provided, a similarity matrix can be naturally constructed based on the node features. The k-means algorithm can be run directly on the given graph structure or on the constructed similarity matrix to obtain clustering results.
Eigenvalue decomposition is introduced through a normalized graph Laplacian matrix, and then, eigenvectors with relatively small eigenvalues are fed into a clustering algorithm to obtain clusters \cite{von2007tutorial}, \cite{newman2006modularity}.
Newman et al. further used Laplacian eigenmaps to group nodes with a higher-than-average density of edges \cite{newman2006finding}.
In addition to eigenvalue decomposition, Girvan et al. proposed a method in which centrality indices are used to locate cluster boundaries \cite{girvan2002community}.
Hastings et al. resorted to the techniques of belief propagation \cite{gallager1962low}, observing that a graph has a low density of loops \cite{hastings2006community}.
By combining both node features and graph structure information, a nonnegative matrix factorization method has also been proposed in which the node attribution matrix is decomposed and the graph structure is utilized to construct regularization terms \cite{cai2008non}, \cite{gu2009co}.
Because DeepWalk is considered equivalent to matrix factorization, Yang et al. proposed to incorporate text features of nodes into network representation learning within the matrix factorization framework \cite{yang2015network}.
Xia et al. also designed a Markov-chain-based method to explicitly handle the possible noise implied in graph structure information and node features \cite{xia2014robust}.

\subsection{Deep Representations for Graph Clustering}
With its powerful representation capabilities, deep learning has been widely applied in the graph domain \cite{8486671}, \cite{ji2020hopgat}.
We intuitively split the deep learning methods used in this field into two main categories --- autoencoders and GCNs \cite{8822591}.
Autoencoders are usually designed as unsupervised learning models. Tian et al. first proposed GraphEncoder, which learns a nonlinear embedding of the original graph by means of stacked autoencoders \cite{tian2014learning}. This method differs from the SVD-based dimension reduction method in which the original representation space is merely projected into a new space with a lower rank through linear projection.
Later, a random surfing model, called deep neural networks for graph representation (DNGR), combined with stacked denoising autoencoders was designed \cite{cao2016deep}.
Another class of graph clustering methods is based on GCNs, in which node representations can be updated by aggregating messages from neighboring vertices. GAEs and VGAEs apply GCNs to encode representations and an inner product decoder for learning \cite{kipfvariational}.
To handle sparse data, second-order proximity (local pairwise second-order similarity within one hop) was exploited in \cite{wang2016structural} to preserve both global and local network structures through a deep structural network embedding method.
By combining autoencoders and GCNs, Wang et al. also proposed a goal-directed deep attentional autoencoder that simultaneously learns the importance of neighboring nodes and soft labels from the graph embedding itself \cite{wang2019attributed}.

Although deep learning techniques have been widely applied, the aforementioned methods can merely aggregate information from neighbors within a limited distance. To address this shortcoming, the SDCN was developed by designing a delivery action to construct connection between an autoencoder and a GCN with three layers. Wang et al. proposed the MGAE approach, which increases the possible number of convolutional layers to 3. AGC was recently proposed to support higher-order graph convolution with a novel low-pass graph filter, which utilizes the smoothness of the graph signals.
Our method, NAS-GC (AS-GC), is closely related to deep approaches of this kind. Concretely, based on the smoothness saturation of attributed graphs and the surrounding environment of each node, we propose a smoothness sensor that can adaptively choose the appropriate order of graph convolution and a fine-grained node-wise mechanism for every vertex, along with an effective self-supervision criterion.

\section{Method}
\subsection{Preliminaries}

\begin{table}[!htbp]
  \renewcommand{\arraystretch}{1.3}
  \caption{Important notations used in this paper}
  \centering
  \begin{tabular}{c|l}
    \hline
    \multicolumn{1}{c|}{Notation} & Short explanation \\
    \hline
      $G$ & Frequency response function of a filter \\
      $G^k X$ & $k$-order graph convolution on the initial feature $X$\\
      \hline
      $s^{k}$ & Smoothness state when $k$-order graph convolution is \\
      & conducted \\
      $p^{\mathcal G}_{st}(\cdot)$ & A state transition (st) model for calculating the \\
      & smoothness of an entire graph ($\mathcal G$) \\
      $q^k_{ss}$ & Smoothness saturation (ss) of $k$-order graph convolution \\
      $N$ & Selected order of graph convolution for the graph \\
      $\overline X$ & Updated representation of all nodes \\
      \hline
      $s^k_i$ & Smoothness state of node $i$ when $k$-order graph convolution \\
      & is conducted \\
      $p^{\mathcal V}_{st}(\cdot)$ & A state transition (st) model for calculating the smoothness \\
      & of an individual node ($\mathcal V$) \\
      $q^{k,i}_{ss}$ & Smoothness saturation (ss) of node $i$ with $k$-order graph \\
      & convolution \\
      $N_i$ & Selected order of graph convolution for node $i$ \\
      $\overline x_i$ & Updated representation of node $i$ \\
    \hline
  \end{tabular}
  \label{notations}
\end{table}

Given a graph $\mathcal{G} = (\mathcal{V}, \mathcal{E}, X)$, $\mathcal{V} = \{v_1, ..., v_n\}$ is the set consisting of all nodes and $\mathcal{E}$ is the edge set which can be represented as an adjacency matrix $A$.
$X$ is the feature matrix $[x_1, ..., x_n]^T \in \mathbb R^{n\times m}$, where $x_i$ denotes the feature representation of vertex $v_i$ and $m$ represents the number of features. Our goal is to assign each node to a cluster. All possible clusters are collected in a set $\mathcal C = \{c_1, ..., c_r\}$, where $r$ is the number of candidate clusters. In addition, we define two operations that will be frequently used in our paper. $[X]_j$ denotes the collection of the $j$-th column in matrix $X$, while $[X]^j$ represents the extraction of the $j$-th row. Table \ref{notations} lists some important notations that will be used throughout the rest of the paper.

\subsection{Smoothness of Graph Signals}
We first introduce some basic notations. Given the adjacency matrix $A$ of graph $\mathcal G$, the degree matrix and the graph Laplacian can be expressed as $D=diag(d_1, ..., d_n)$ and $L=D-A$, respectively. This Laplacian can be eigen-decomposed as $U \Lambda U^{-1}$, where $\Lambda=diag(\lambda_1, ..., \lambda_n)$ is a diagonal matrix of the eigenvalues and $U=[u_1, ..., u_n]$ is the matrix of the corresponding eigenvectors. Moreover, the symmetrically normalized graph Laplacian is $L_s = I - D^{-\frac{1}{2}}AD^{-\frac{1}{2}}$, which can also be eigen-decomposed in the same way as $L$.

A \textbf{graph signal} can be represented as a vector $ f= [f(1),\cdots,f(n)]$, where $f:\mathcal V \rightarrow \mathbb R$ is a real-valued function on the nodes of a graph. The input feature matrix $X \in \mathbb R^{n \times m}$ can be split into $m$ individual graph signals, where column $[X]_j$ corresponds to the $j$-th signal. Then, each graph signal can be decomposed into a linear combination of the eigenvectors \cite{shuman2013emerging}:
\begin{equation}
  [X]_j = \sum_{i=1}^n e_i u_i
\end{equation}
where $e_i$ is the coefficient of $u_i$ for the $j$-th graph signal and is proportional to the strength of the basis signal $u_i$.

The smoothness assesses the degree of similarity among the feature representations of nearby nodes in a graph. Specifically, the smoothness of a basis signal can be calculated using the Laplacian-Beltrami operator \cite{chung1997spectral}:

\begin{equation}
  \Omega(u_q) = \frac{1}{2} \sum_{(v_i, v_j)\in \mathcal E} {a_{i,j} \Vert \frac{u_q^i} {\sqrt {d_i}} - \frac{u_q^j}{\sqrt {d_j}} \Vert^2_2 } = u_q^T L_s u_q = \lambda_q
\end{equation}

where $u_q^i$ denotes the $i$-th element of eigenvector $u_q$ and $a_{i,j}$ is the element located in the $i$-th row and $j$-th column of adjacency matrix $A$. It can be observed that the smoothness of a basis signal is equivalent to the corresponding eigenvalue, with smaller eigenvalues indicating smoother basis signals.
We thus refer to signals with small eigenvalues as low-frequency (smoother) signals.

\subsection{Node-wise Adaptive Smoothness-transition Graph Convolution}

In this section, we seek a graph convolutional approach to obtain better graph representations for clustering.
We first propose a naive but intuitive prototype and gradually evolve it into its mature version, NAS-GC. The entire framework of NAS-GC is illustrated in Fig. \ref{overall_framework}.

\subsubsection{Evolutionary Process of the Overall Objective of the Smoothness Sensor}
\

\noindent\textbf{$\bullet$ Order-fixed Graph Convolution}

Smooth graph signals tend to cause nearby nodes to have similar representations, which is consistent with the way a cluster in a graph tends to be composed of adjacent nodes. Lower-frequency signals correspond to smoother graph signals and thus help to form the nodes in an attributed graph into clusters more easily.

A low-pass graph filter is a function for producing low-frequency basis signals from relatively high-frequency signals for various downstream tasks. We first define a frequency response function for a filter as follows: $G = Up(\Lambda)U^{-1}$, where  $p(\Lambda)=diag(p(\lambda_1), ..., p(\lambda_n))$.

Given a specified graph filter, we can execute a first-order graph convolution as follows:
\begin{equation}
  [\overline X]_j = [GX]_j = [Up(\lambda)U^{-1}X]_j=\sum_{i=1}^n p(\lambda_i)e_i u_i
\end{equation}
where $[\overline X]_j$ is the filtered version of the $j$-th graph signal from $X$. $p(\lambda_i)$ is assigned to preserve low-frequency basis signals and remove high-frequency ones by scaling the values of $e_i$.

When a first-order convolution is conducted, the representation of every node is updated by aggregating its 1-hop neighbors. In this way, the information from long-distance neighbors is discarded, which could lead to severe problems in a large but highly sparse graph, resulting in an undersmooth graph signal. To alleviate this problem, the concept of $k$-order graph convolution is introduced.

We formulate the $k$-order graph convolution process as follows:
\begin{equation}
  [\overline X]_j = [G^kX]_j = [Up(\lambda)^kU^{-1}X]_j
\end{equation}

Finally, with a predefined $k$, the filtered graph signals can be fed to a downstream algorithm to perform clustering. $[G^kX]_j$ represents the $j$-th filtered graph signal under $k$-order convolution.

\noindent\textbf{$\bullet$ Adaptive Graph Convolution}

Oversmooth graph signals can exert a clear negative effect on clustering by causing two vertices that should belong to distinct clusters to possess similar representations with limited discriminability.
To achieve a balance between undersmoothness and oversmoothness, the selection of a suitable $k$ for the $k$-order convolution process is naturally critical for learning effective representations of the graph and, consequently, for ensuring good performance in downstream tasks, e.g., attributed graph clustering.

A naive paradigm for solving this problem is to adopt a score function or probability maximization model. We first define a maximum order $K$. Then, graph signals filtered by graph convolutions of different orders are fed into this model one by one, and the corresponding levels of smoothness saturation are evaluated. We can choose the order with the maximum smoothness saturation as the optimal order for graph convolution, as follows:
\begin{equation}
  \label{MotivationofOverallObject:E1}
  N = \mathop{\arg\max}_{k=1,...,K} p_{agc}(G^k X)
\end{equation}
where $p_{agc}(\cdot)$ denotes a score function or probabilistic model to output the value of smoothness saturation associated with $k$-order graph convolution.

\noindent\textbf{$\bullet$ Adaptive Smoothness-transition Graph Convolution}

The above solution simply assumes that there is no relationship among the filtered graph signals obtained under different orders of graph convolution.
Considering that the low-pass frequency response function $p(\cdot)$ should be constantly nonincreasing and nonnegative for all input, the smoothness of the filtered node features will monotonically increase with increasing $k$ \cite{zhang2019attributed}. This can be expressed as follows:
\begin{equation}
  \Omega(\frac{[G^k X]_j}{||[G^k X]_j||_2}) \leq \Omega(\frac{[G^{k-1} X]_j}{||[G^{k-1} X]_j||_2})
\end{equation}
where $||\cdot||_2$ is used to project different graph signals to a common scale.

Thus, we can theoretically consider graph convolutions of increasing order as a process of a gradual change in smoothness. Inspired by this observation, we convert the aforementioned smoothness saturation objective Equation (\ref{MotivationofOverallObject:E1}) into a sequence prediction problem, in which the smoothness saturation associated with the current order of graph convolution depends on the previously filtered signals. We formulate a state transition model $p_{st}^{\mathcal{G}}(\cdot)$ for the entire graph as follows:
\begin{equation}
  \label{MotivationofOverallObject:E2-1}
  q_{ss}^k = p_{st}^{\mathcal{G}}(s^{k-1}, G^k X)
\end{equation}
\begin{equation}
  \label{MotivationofOverallObject:E2}
  N = \mathop{\arg\max}_{k=1,...,K} q_{ss}^k
\end{equation}
where $s^{k-1}$ denotes the smoothness state of the entire graph resulting from the last graph convolution.

Once $N$ is fixed, we can achieve the final representation of the nodes through $N$-order graph convolution:
\begin{equation}
  \label{MotivationofOverallObject:E5-1}
  \overline X = G^N X
\end{equation}

Considering the possible variance caused by noise from some of the filtered graph signals during the convolution procedure, we further propose a linear composition of multiple graph convolutions of distinct orders.
We represent the accumulated value of smoothness saturation with respect to a threshold $\epsilon$.
The ultimate form of our task objective can be written as follows:
\begin{equation}
  \label{MotivationofOverallObject:E8-1}
  N = min \{ k': \sum_{k=1}^{k'} q_{ss}^k \geq \epsilon \}
\end{equation}
\begin{equation}
  \label{MotivationofOverallObject:E7-1}
  \overline X = \sum_{k=1}^N q_{ss}^k \cdot G^k X
\end{equation}
where $q_{ss}^k$ is accumulated to $\epsilon$.

\noindent\textbf{$\bullet$ Node-wise Adaptive Smoothness-transition Graph Convolution}

Notably, in the above solutions, the optimal order $N$ of graph convolution is determined only by the global smoothness saturation of the entire graph.
In $k$-order graph convolution, the node representations are updated by iteratively aggregating the features of all $k$-hop neighbors.
However, the smoothness saturation of different nodes subjected to convolution of the same order $N$ could differ considerably.
Intuitively, the density of the nodes can vary greatly in attributed graphs. A relatively isolated node with few neighbors could require a larger $k$ to introduce more distant nodes to obtain more information, while a node with more neighbors usually can efficiently gather information in fewer hops.
Thereafter, we should consider the surrounding environment of every node and adjust the optimal order $N$ of graph convolution separately for each node to guarantee the collection of sufficient information from within an appropriate distance without introducing irrelevant noise.
Accordingly, we further optimize our objective Equation (\ref{MotivationofOverallObject:E2-1}) in a more fine-grained, node-wise manner:

\begin{equation}
  \label{MotivationofOverallObject:E2-11}
 q_{ss}^{k, i} = p_{st}^{\mathcal{V}}(s_i^{k-1}, [G^k X]^i)
\end{equation}
where the smoothness state of node $v_i$ resulting from $k$-order graph convolution is denoted by $s_i^{k-1}$ and $p_{st}^{\mathcal{V}}(\cdot)$ is the corresponding node-wise state transition function.

Driven by a motivation similar to that for Equation (\ref{MotivationofOverallObject:E7-1}), to reduce the extra fluctuations introduced by different nodes, we accumulate the node representation of node $v_i$ produced by graph convolutions of distinct orders $k$, combined with the corresponding smoothness saturation.
We formulate this process as follows:
\begin{equation}
  \label{MotivationofOverallObject:E8}
  N_i = min \{ k': \sum_{k=1}^{k'} q_{ss}^{k, i} \geq \epsilon \}
\end{equation}
\begin{equation}
  \label{MotivationofOverallObject:E7}
  \overline x_i = \sum_{k=1}^{N_i} q_{ss}^{k, i} \cdot [G^k X]^i
\end{equation}

\subsubsection{Internal Structure of the Smoothness Sensor}

With the established overall objective of customizing the node-wise order of graph convolution in Equation (\ref{MotivationofOverallObject:E7}), in this section, we illustrate the internal structure of our proposed NAS-GC and AS-GC methods. We build our methods upon an adaptive computation time mechanism that has been proposed for RNNs \cite{graves2016adaptive}. We first present the details of realizing adaptive graph convolution at the graph level, as in Equation (\ref{MotivationofOverallObject:E7-1}), and then extend the operations to the node level, as in Equation (\ref{MotivationofOverallObject:E7}).

To compute the accumulated smoothness saturation, the remaining challenge is to design an effective smoothness-transition model, as in Equation (\ref{MotivationofOverallObject:E2-1}), and adapt it to the graph domain with graph convolution.

First, we choose a low-pass graph filter such as that in \cite{zhang2019attributed}. The corresponding frequency response function is:
\begin{equation}
  \label{Model:E1}
  p(\lambda_q) = 1-\frac{1}{2}\lambda_q
\end{equation}
where $\lambda_q$ is the $q$-th eigenvalue derived from the symmetrically normalized graph Laplacian.
Accordingly, the filtered graph signals under $k$-order convolution can be specified as follows:
\begin{equation}
  \label{Model:E2-1}
  G^k X = U(I-\frac{1}{2}\Lambda)^kU^{-1}X
\end{equation}

We model the gradual smoothing process as a sequence in which the smoothness under $k$-order graph convolution depends on the historical smoothness states, i.e., $1$, ..., $k-1$. A variable $s^k$ records the state under $k$-order graph convolutions:
\begin{equation}
  \label{Model:E3}
  s^k=
  \begin{cases}
    \mathcal S(0, g(X)), & \text{if}\ k = 0 \\
    \mathcal S(s^{k-1}, g(G^k X)), & \text{otherwise}
  \end{cases}
\end{equation}
where $X$ is the original feature matrix of the graph and $G^k X$ represents the filtered signals under $k$-order graph convolution.
The function $g(\cdot)$ is a neural network projecting $\mathbb R^{n\times m}$ to $\mathbb R^{m}$.
The model $\mathcal S(\cdot)$ can be implemented with RNNs \cite{hochreiter1997long} and GRUs \cite{cho2014learning}, in which the first and second parameters correspond to the previous state under $(k-1)$-order graph convolution and the current filtered graph signal. In particular, the state in the initial step is set to a zero vector.

Once the current state is recorded, we introduce an extra unit to sense whether the smoothness is already saturated:
\begin{equation}
  \label{Model:E4}
  h^k = \sigma (W_h s^k + b_h)
\end{equation}
where $W_h$ and $b_h$ represent the trainable parameters and bias, respectively. $\sigma(\cdot)$ denotes the sigmoid function, which projects the output to a fixed range of $(0, 1)$. We then accumulate the estimated $\{h^1, h^2,...\}$. Once the accumulated value exceeds the threshold, the smoothness is saturated and $N$ is achieved:
\begin{equation}
  \label{Model:E5}
  N = min\{M, min \{ k': \sum_{k=1}^{k'} h^k \geq \epsilon \}\}
\end{equation}
where $M$ is a hyperparameter to limit the maximum order of graph convolution.

To guarantee that the accumulated value reaches exactly $\epsilon$, we specially address the N-order graph convolution.
The complete calculation of the smoothness saturation can be written as follows:
\begin{equation}
  \label{Model:E7}
  q_{ss}^k=
  \begin{cases}
    \epsilon - \sum_{k=1}^{N-1}h^k, & \text{if}\ n=N \\
    h^k, & \text{otherwise}
  \end{cases}
\end{equation}

Thus, the smoothness-transition model is complete. Finally, an updated representation can be obtained via Equation (\ref{MotivationofOverallObject:E7-1}). The graph convolution based on the accumulated smoothness-transition model at the graph level is achieved. We can similarly adopt the aforementioned operations at the node level.

Given a specific node, a variable $s^k_i$ is assigned to replace $s^k$ for recording the smoothness state of node $v_i$ under $k$-order graph convolution:
\begin{equation}
  \label{Model:E3-1}
  s^k_i=
  \begin{cases}
    \mathcal S(0, x_i), & \text{if}\ k = 0 \\
    \mathcal S(s_i^{k-1}, [G^k X]^i), & \text{otherwise}
  \end{cases}
\end{equation}
where $x_i$ is the original feature of node $v_i$.
Similarly, $h^k$ is transformed into $h^k_i$, which is associated with node $v_i$:
\begin{equation}
  \label{Model:E4}
  h^k_i = \sigma (W_h s^k_i + b_h)
\end{equation}

By accumulating the estimated values $\{h_i^1, h_i^2,...\}$,
the boundary of smoothness saturation can be drawn as follows:
\begin{equation}
  \label{Model:E5}
  N_i = min\{M, min \{ k': \sum_{k=1}^{k'} h_i^k \geq \epsilon \}\}
\end{equation}

Finally, we utilize the maximum accumulated value $\epsilon$ to obtain the ultimate smoothness saturation of node $v_i$ under $k$-order graph convolution:
\begin{equation}
  \label{Model:E7-1}
  q_{ss}^{k,i}=
  \begin{cases}
    \epsilon - \sum_{k=1}^{N_i-1}h_i^k, & \text{if}\ n=N_i \\
    h^k_i, & \text{otherwise}
  \end{cases}
\end{equation}

Thus, the node-wise adaptive smoothness-transition model becomes approachable.

\subsection{Self-supervised Clustering Learning}
Although NAS-GC (AS-GC) has the potential to enable graph convolution that can yield effective node representations for downstream clustering tasks, a supervision mechanism to guide the training process of the proposed model for a typical unsupervised clustering task is still lacking.
Furthermore, it should be noted that NAS-GC provides a local perspective for evaluating smoothness in which the smoothness of each node is assessed in accordance with the surrounding environment within a relatively limited radius, but without any potential cluster-related information for the node. To be concrete, the global outlook --- the distribution of the nodes with respect to clusters --- is neglected, which leads to that node pairs separated by a long distance may nevertheless belong to the same cluster, whereas pairs located nearby may belong to different clusters. Based on this intuition, we propose and illustrate a self-supervision strategy that fills in the gap between the local and global perspectives and thus empowers adaptability with respect to any graph structures. In addition, this learning tactic is also available for AS-GC.

Given the learned features $\overline X = [\overline x_1, ..., \overline x_n]^T$, we apply a linear kernel $K=\overline X \overline X^T$ to quantify the similarity of node pairs. Then, we use $W=\frac{1}{2}(|K|+|K^T|)$ to make $K$ symmetric and nonnegative \cite{nikolentzos2017matching}. The function $|\cdot|$ represents taking the absolute value of each element in the matrix. Once the similarity matrix has been obtained, the eigenvectors associated with the $r$ largest eigenvalues of $W$, where $r$ is the number of expected clusters, are calculated and passed to the k-means algorithm to obtain the ultimate cluster partitions. However, to more effectively perform the clustering task, the following factors should be considered.

\textbf{Intracluster tightness.} A good cluster partition should have a small intracluster distance.
It is natural to apply the indicator of tightness, which is the average length of all lines in $C(i)$ connected to node $v_i$ \cite{rousseeuw1989graphical}:
\begin{equation}
  \label{ClusteringTheory:E1}
  tig(i) = \frac{\sum_{v_j\in C(i)}dis(i, j)}{|C(i)|}
\end{equation}
where $C(i)$ denotes the node set belonging to the cluster to which vertex $v_i$ is assigned and $dis(\cdot)$ is a function for measuring the dissimilarity between two objects. Benefiting from this global measurement, the representations of two nodes situated far from each other but belonging to the same cluster can be detected and adjusted.
We then extend $tig(i)$ to the entire graph and adopt it as part of our loss function, denoted by $\mathcal L_{tig}$:
\begin{equation}
  \label{ClusteringTheory:E4}
  \mathcal L_{tig} = \frac{1}{|\mathcal C|}\sum_{c\in\mathcal C}\frac{1}{|c|(|c|-1)}\sum_{v_i, v_j \in c, v_i\neq v_j}||\overline x_i - \overline x_j||_2
\end{equation}

\textbf{Intercluster separation.} A good cluster partition should have a large intercluster distance. Compared with the tightness, although the separation between clusters also plays a critical role, the intercluster separation is unfortunately neglected by other methods. For instance, in AGC \cite{zhang2019attributed}, the node features become smoother as the order $k$ of graph convolution increases, which will ultimately reduce both the intracluster and intercluster distances. Based on our proposed algorithm, however, intercluster separation can be equally considered in the opposite direction --- the separation is expected to be large, while the tightness should be small.
We formally define the inter-cluster separation as follows:

\begin{equation}
  \label{ClusteringTheory:E2}
  sep(i) = \frac {\sum_{c\in C'(i)} \frac{\sum_{v_j\in c}dis(i, j)}{|c|}} {|C'(i)|}
\end{equation}
where $C'(i)$ represents the set of nodes belonging to a different cluster than that to which the vertex $v_i$ belongs. Similar to the benefit offered by the intracluster tightness, the representation of nodes located nearby but assigned to different clusters are highlighted under this intercluster separation indicator.
We also adopt this indicator as part of our loss function, denoted by $\mathcal L_{sep}$:
\begin{equation}
  \label{ClusteringTheory:E5}
  \mathcal L_{sep} = \frac{1}{|\mathcal C|}\sum_{c\in\mathcal C}\frac{1}{|c|(|c|-1)}\sum_{v_i \in c, v_j\notin c }||\overline x_i - \overline x_j||_2
\end{equation}

\textbf{Trade-off between tightness and separation.}
Some bias between tightness and separation is inevitable with respect to the number of expected clusters and the distribution of nodes in the graph; this is the previously mentioned global structure information. From this global perspective, local observations can be compensated.
To be precise, this adaptivity is empowered by the fact that the trade-off between tightness and separation can be adjusted with respect to different graph structures.
We consider the tightness and separation across all nodes for a cluster partition and combine them into an overall expression:
\begin{equation}
  \label{ClusteringTheory:E6}
  \mathcal L = \lambda_{tig} \mathcal L_{tig} + \lambda_{sep} \frac{1}{\mathcal L_{sep}}
\end{equation}
where a larger $\lambda_{tig}$ drives the vertices within a cluster to be tighter and $\lambda_{sep}$ drives the nodes to be well separated between clusters.
$\lambda_{tig}$ and $\lambda_{sep}$ are adversarial parameters used to control the tradeoff between these two indicators and adapt the method to distinct graph networks.

Considerable effort could be required to determine the optimal parameters, $\lambda_{tig}$ and $\lambda_{sep}$, through a comprehensive analysis of the nature of the data of interest. Therefore, we propose a practical and automatic selection strategy for seeking the optimal parameters for various datasets.

Instead of directly tuning the hyperparameters $\lambda_{tig}$ and $\lambda_{sep}$ by exploring the dataset characteristics, we start by observing the proportion of $\mathcal L_{tig}$ with respect to $\frac{1}{\mathcal L_{sep}}$, which can be roughly approximated by the value obtained after executing the first epoch. Thereafter, we can balance the two terms $\lambda_{tig} \mathcal L_{tig}$ and $\lambda_{sep} \frac{1}{\mathcal L_{sep}}$ in Equation (\ref{ClusteringTheory:E6}).
We conduct a grid search on the proportion sequence, with the proportion between the two terms ranging from $1:3$ to $1:50$. Then, a favorable choice can be automatically revealed.

\section{Experiments}
\subsection{Data}

\begin{table}[!htbp]
  \renewcommand{\arraystretch}{1.3}
  \caption{Basis statistics of datasets}
  \centering
  \begin{tabular}{ccccc}
    \hline
    Dataset & \#Nodes & \#Edges & \#Features & \#Classes \\
    \hline
    Cora & 2,708 & 5,429 & 1,433 & 7 \\
    Citeseer & 3,327 & 4,732 & 3,703 & 6 \\
    Pubmed & 19,717 & 44,338 & 500& 3 \\
    Wiki & 2,405 & 17,981 & 4,973 & 17 \\
    \hline
    \end{tabular}
  \label{basis_statistics}
\end{table}

We apply four datasets as our benchmark datasets. Cora, Citeseer and Pubmed can be characterized as citation networks in which nodes represent various documents and edges connect two documents with a citation relation. Each document is classified into a particular class. The last dataset, Wiki, is a webpage network in which webpages (nodes) are connected by page link relations (edges). In Cora and Citeseer, the initial node features are represented through the bag-of-words approach, while in Pubmed and Wiki, tf-idf word vectors are used. The details of the numbers of nodes, edges, features and classes are listed in Table \ref{basis_statistics}.


\subsection{Baselines and Evaluation Metrics}
To evaluate the performance of the proposed method, 12 state-of-the-art methods are used as baseline methods, which can be grouped into three categories according to their inputs:
\begin{itemize}
  \item Node-feature-based methods. Similarity matrices are first constructed from the input node representations, and clustering techniques are then conducted on the constructed matrices. Typical examples are k-means \cite{hartigan1979algorithm} and spectral clustering (spectral-f).
  \item Graph-structure-based methods. Adjacency matrices and other graph representations are mainly considered, such as DeepWalk \cite{perozzi2014deepwalk}, DNGR \cite{cao2016deep}, and structure-based spectral clustering (spectral-g).
  \item Methods based on both node features and graph structures. Deep graph neural networks are employed to combine both kinds of elements in the GAE, VGAE, MGAE, SDCN, AGC, adversarially regularized graph autoencoder (ARGE) and adversarially regularized variational graph autoencoder (ARVGE) \cite{pan2018adversarially} approaches.
\end{itemize}

The parameter settings of the AGC \footnote{\url{https://github.com/karenlatong/AGC-master}} and SDCN \footnote{\url{https://github.com/bdy9527/SDCN}} methods are consistent with their published codes. The implementations of the other baselines are inherited from their original papers and several parameters are optimized as mentioned in \cite{zhang2019attributed}. We apply three popular cluster evaluation metrics \cite{zhang2019attributed}, \cite{aggarwal2014data}: clustering accuracy (Acc), normalized mutual information (NMI) and macro F1-score (F1).

\subsection{Implementation Details}
\label{section:ImplementationDetails}
For NAS-GC, the maximum order of graph convolution is set to 40 for Cora, Citeseer and Wiki and to 120 for Pubmed. GRUs with a hidden size of 200 are chosen as the function $\mathcal S(\cdot)$ for Cora, Citeseer and Wiki. We uniformly set the smoothness saturation to 1 for all datasets. RNNs equipped with 50 hidden units are employed for Pubmed.
We train all models with the Adam optimizer \cite{kingma2014adam}. The learning rate is 0.01 for Cora, 0.005 for Pubmed, 0.003 for Citeseer, and 0.0001 for Wiki. For Wiki, the learning rate is also annealed by 0.96 when the epoch index is larger than 10.
The bias hyperparameter $\lambda_{tig}$ is uniformly set to 1 for all datasets, while $\lambda_{sep}$ is chosen to be 50, 350, 0.0005 and 10 for Cora, Citeseer, Pubmed and Wiki, respectively. These bias hyperparameters are automatically obtained via our proposed automatic selection strategy for self-supervised clustering learning. In addition, we will present a detailed analysis of this hyperparameter later.

The implementation details of AS-GC are similar to those of NAS-GC, with the exception that the maximum order of graph convolution is set to 25 and 30 for the Cora and Wiki datasets, respectively, and the learning rate for Citeseer is 0.03.

We uniformly customize an early termination mechanism such that if the standard deviation of the losses produced in the last 5 epochs is smaller than 0.001 (NAS-GC) or 0.1 (AS-GC), the training process will be terminated in advance. This mechanism is applied on all datasets, with a maximum of 200 epochs.

\subsection{Results}

\begin{table*}[!htbp]
  \renewcommand{\arraystretch}{1.3}
  \caption{Performance on clustering tasks}
  \centering
  \begin{tabular}{lc|ccc|ccc|ccc|ccc}
    \hline
    \multicolumn{1}{c}{\multirow{2}{*}{Category}} & \multicolumn{1}{c|}{\multirow{2}{*}{Method}} & \multicolumn{3}{c|}{Cora} & \multicolumn{3}{c|}{Citeseer} & \multicolumn{3}{c|}{Pubmed} & \multicolumn{3}{c}{Wiki}\\
  \multicolumn{2}{c|}{}&Acc\%&NMI\%&F1\%&Acc\%&NMI\%&F1\%&Acc\%&NMI\%&F1\%&Acc\%&NMI\%&F1\% \\
    \hline
    \multicolumn{1}{c}{\multirow{2}{*}{Node feature}}
    & K-means & 34.65 & 16.73 & 25.42 & 38.49 & 17.02 & 30.47 & 57.32 & 29.12 & 57.35 & 33.37 & 30.20 & 24.51 \\
    & Spectral-f & 36.26 & 15.09 & 25.64 & 46.23 & 21.19 & 33.7 & 59.91 & 32.55 & 58.61 & 41.28 & 43.99 & 25.20 \\
    \hline
    \multicolumn{1}{c}{\multirow{3}{*}{Graph structure}}
    & Spectral-g & 34.19 & 19.49 & 30.17 & 25.91 & 11.84 & 29.48 & 39.74 & 3.46 & 51.97 & 23.58 & 19.28 & 17.21 \\
    & DeepWalk & 46.74 & 31.75 & 38.06 & 36.15 & 9.66 & 26.70 & 61.86 & 16.71 & 47.06 & 38.46 & 32.38 & 25.74 \\
    & DNGR & 49.24 & 37.29 & 37.29 & 32.59 & 18.02 & 44.19 & 45.35 & 15.38 & 17.90 & 37.58 & 35.85 & 25.38 \\
    \hline
    \multicolumn{1}{c}{\multirow{7}{*}{Both}}
    & GAE & 53.25 & 40.69& 41.97 & 41.26 & 18.34 & 29.13 & 64.08 & 22.97 & 49.26 & 17.33 & 11.93 & 15.35 \\
    & VGAE & 55.95 & 38.45 & 41.50 & 44.38 & 22.71 & 31.88 & 65.48 & 25.09 & 50.95 & 28.67 & 30.28 & 20.49 \\
    & MGAE & 63.43 & 45.57 & 38.01 & 63.56 & 39.75 & 39.49 & 43.88 & 8.16 & 41.98 & 50.14 & 47.97 & 39.20 \\
    & ARGE & 64.00 & 44.90 & 61.90 & 57.30 & 35.00 & 54.60 & 59.12 & 23.17 & 58.41 & 41.40 & 39.50 & 38.27 \\
    & ARVGE & 63.80 & 45.00 & 62.70 & 54.40 & 26.10 & 52.90 & 58.22 & 20.62 & 23.04 & 41.55 & 40.01 & 37.80 \\
    & SDCN & 62.22 & 35.16 & 57.73 & 65.96 & 38.71 & 63.62 & 65.20 & 28.03 & 65.64 & 42.20 & 37.79 & 24.05 \\
    & AGC & 68.92 & 53.68 & 65.61 & 67.00 & 41.13 & 62.48 & 69.78 & 31.59 & 68.72 & 47.65 & 45.28 & 40.36 \\
    \hline
    \multicolumn{1}{c}{\multirow{2}{*}{Ours}}
    & AS-GC & 69.45 & 54.38 & 66.06 & 68.49 & 42.49 & 63.83 & 69.90 & 32.78 & 69.07 & 51.16 & 47.70 & 42.14 \\
    & NAS-GC & \textbf{71.94} & \textbf{54.89} & \textbf{69.01} & \textbf{69.58} & \textbf{43.78} & \textbf{64.73} & \textbf{70.7} & \textbf{36.79} & \textbf{70.40} & \textbf{56.24} & \textbf{51.07} & \textbf{44.79} \\
    \hline
    \end{tabular}
  \label{performance}
\begin{tablenotes}
\item[1] \textbf{note:} Bold type indicates the best scores.
\end{tablenotes}
\end{table*}

Table \ref{performance} records the average performance over 10 repetitions of the clustering task on each dataset.

Comparing the node-feature-based methods with the graph-structure-based methods, we can observe that there is no clear superiority between these two approaches. The node-feature-based spectral-f method performs better on Citeseer and Wiki, while the graph-structure-based DNGR method works better on Cora. On Pubmed, DeepWalk achieves better ACC performance but worse NMI and F1 performance than spectral-f. Therefore, we cannot draw a clear conclusion regarding the superiority of either node- or graph-based methods.

In comparison with the baselines utilizing either node features or graph structure information alone, we note a significant improvement in the 3rd group of methods, which combine these two kinds of information. On Cora, all of these methods outperform the methods in the first two groups by a considerable margin. As a reference, we average the scores of the methods in the first two groups and those of the methods in the third group. Compared to the first two groups, the maximum improvement of the third group is by 21.44\% (Acc), 19.28\% (NMI) and 21.46\% (F1) on Cora, and the minimum improvement is on Wiki, with improvements of 3.57\%, 3.77\% and 7.18\%, respectively. It can be concluded that effectively combining node features and graph structure information can be beneficial for the node representations used for the downstream clustering task.

In terms of all metrics, our proposed methods, both NAS-GC and AS-GC, achieve the best results across all datasets. In particular, we compare NAS-GC with the best performance of the baseline AGC. NAS-GC outperforms AGC by 8.59\% (Acc), 5.79\% (NMI) and 4.43\% (F1) on Wiki, and the average improvements across all 4 datasets are 3.78\% (Acc), 3.71\% (NMI), and 2.94\% (F1). AS-GC also performs better than AGC but is worse than NAS-GC, which is consistent with our core intuition that considering a gradual change in smoothness can provide a means of detecting the smoothness saturation boundary, and node-wise smoothness sensing can further enhance the effectiveness. Below, we will present an extensive study to further explain these results.

In the 3rd group, the numbers of convolutional layers applied differ among the different methods. The GAE, VGAE, ARGE and ARVGAE methods all employ 2nd-order convolutions, while the MGAE and SDCN methods rely on 3rd-order convolutions; hence, these convolutions are relatively shallow. In contrast, AGC performs 12th-, 55th-, 60th- and 8th-order graph convolutions on Cora, Citeseer, Pubmed and Wiki, respectively; these convolutions are much deeper than those of the other baselines and achieve better performance.
The maximum order of graph convolution in our proposed method NAS-GC is 40 for Cora, Citeseer and Wiki and 120 for Pubmed. We will analyze the impact of the order of graph convolution in detail in the next section.

\subsection{Extensive Study}
In this section, we will present several extensive experiments to examine how the proposed methods NAS-GC and AS-GC operate.

\subsubsection{Does the performance of NAS-GC merely depend on higher orders of graph convolution?}

\begin{table*}[!htbp]
  \renewcommand{\arraystretch}{1.3}
  \caption{Performance of NAS-GC with different maximum orders of graph convolution}
  \centering
  \begin{tabular}{c|ccc|ccc|ccc|ccc}
    \hline
    \multicolumn{1}{c|}{\multirow{2}{*}{Method}}& \multicolumn{3}{c|}{Cora} & \multicolumn{3}{c|}{Citeseer} & \multicolumn{3}{c|}{Pubmed} & \multicolumn{3}{c}{Wiki}\\
  \multicolumn{1}{c|}{}&Acc\%&NMI\%&F1\%&Acc\%&NMI\%&F1\%&Acc\%&NMI\%&F1\%&Acc\%&NMI\%&F1\% \\
    \hline
    AGC & 68.92 & 53.68 & 65.61 & 67.00 & 41.13 & 62.48 & 69.78 & 31.59 & 68.72 & 47.65 & 45.28 & 40.36 \\
    \hline
    M=2 & 60.83 & 39.75 & 60.6 & 62.33 & 36.47 & 58.97 & 61.21 & 34.21 & 59.96 & 45.15 & 43.35 & 34.68 \\
    M=5 & 65.03 & 49.79 & 59.91 & 66.66 & 40.70 & 62.23 & 61.76 & 32.49 & 60.49 & \textbf{48.83} & \textbf{48.89} & \textbf{40.77} \\
    M=7 & 65.36 & 49.62 & 60.02 & \textbf{67.59} & \textbf{41.62} & \textbf{63.08} & 62.3 & 31.73 & 61.13 & - & - & - \\
    M=10 & \textbf{70.22} & \textbf{54.10} & \textbf{67.44} & - & - & - & 63.21 & 31.19 & 62.20 & - & - & - \\
    M=20 & - & - & - & - & - & - & 65.94 & 31.93 & 65.27 & - & - & - \\
    M=30 & - & - & - & - & - & - & 68.03 & 33.35 & 67.46 & - & - & - \\
    M=40 & - & - & - & - & - & - & 69.06 & 34.09 & 68.55 & - & - & - \\
    M=50 & - & - & - & - & - & - & 69.69 & 34.85 & 69.12 & - & - & - \\
    M=60 & - & - & - & - & - & - & \textbf{70.26} & \textbf{34.50} & \textbf{69.66} & - & - & - \\
    \hline
    \end{tabular}
  \label{performance_limit_conv}
\begin{tablenotes}
\item[1] \textbf{note:} Bold type indicates the best scores.
\end{tablenotes}
\end{table*}

Although our method significantly outperforms all other baselines via relatively high orders of graph convolution, there is a concern of whether the performance of our approach merely depends on performing convolutions of higher order. To eliminate this concern, we conduct a series of experiments to witness the performance changes as the order of graph convolution gradually increases. We follow the same hyperparameter settings as in the above evaluation test except for the order of graph convolution. We continuously increase the magnitude of the maximum order of graph convolution until the results outperform AGC which almost achieves the best performance across all datasets and metrics. The results are recorded in Table \ref{performance_limit_conv}.

Our model performs better than AGC across all 4 datasets. To be precise, when 10th-, 7th- and 5th-order graph convolution is applied, the proposed method already outperforms AGC with 12th-, 55th- and 8th-order graph convolution in the Cora, Citeseer and Wiki dataset, respectively.
This experiment proves that the superiority of our model does not completely rely on higher orders of graph convolution.

\subsubsection{Does the node-wise mechanism contribute to the performance of NAS-GC?}
We have proven that deeper graph convolution operations are not the sole reason that NAS-GC can significantly outperform the considered baselines. Another possible influential factor could be the node-wise mechanism, which is based on the premise that the order of graph convolution should be chosen for each node individually in accordance with its concrete surroundings. Although the superiority of this approach has already been verified through performance experiments in which NAS-GC outperforms AS-GC across all datasets, we further present an extensive investigation of the node-wise order of graph convolution, the variable $N_i$, to more comprehensively illustrate this point.

\begin{figure}
 \centering
 \includegraphics[scale=0.32]{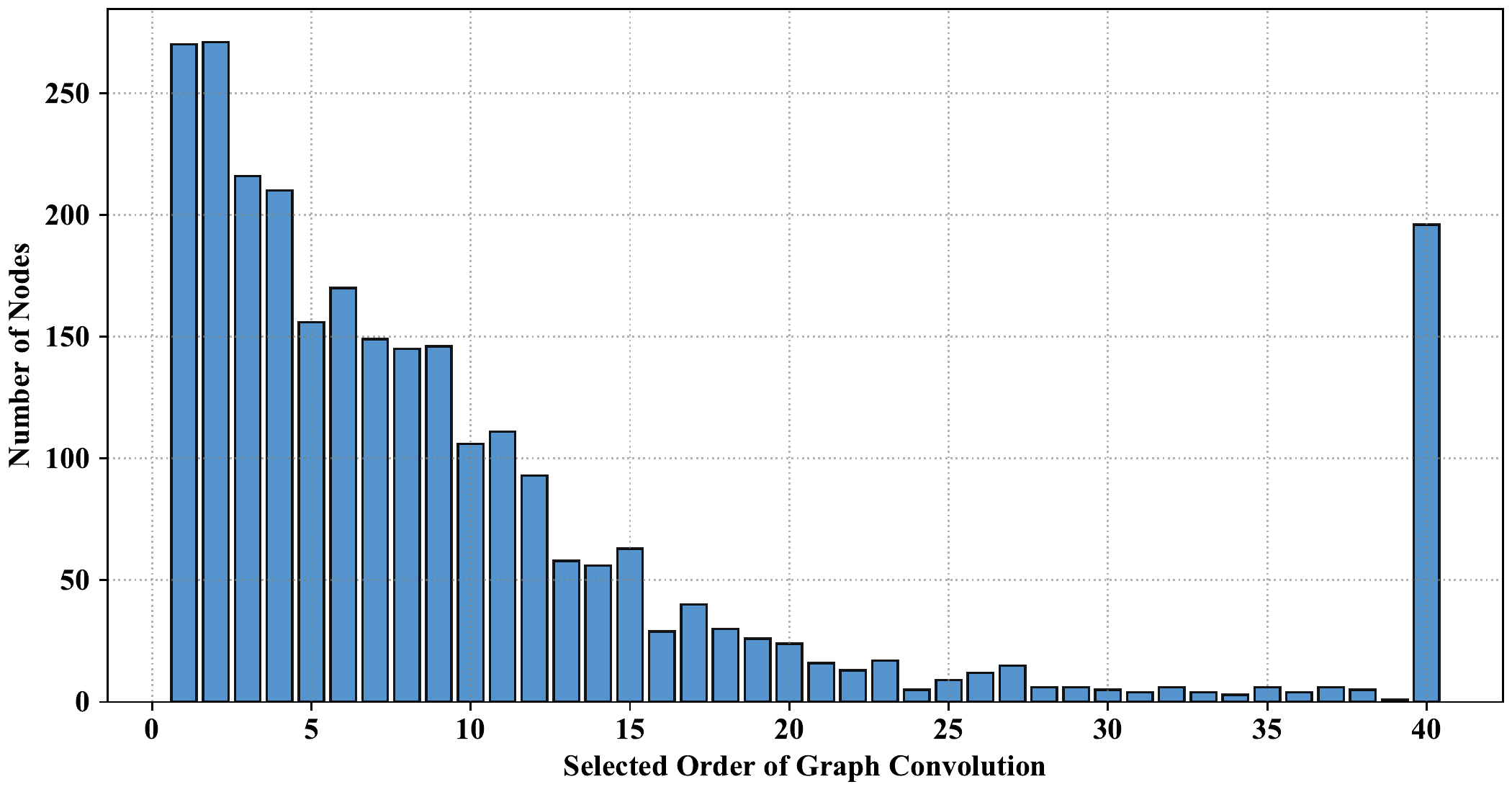}
 \caption{Number of vertices with respect to the selected order of graph convolution for the Cora dataset.}
 \label{distribution_order_cora}
\end{figure}
\begin{figure}
 \centering
 \includegraphics[scale=0.32]{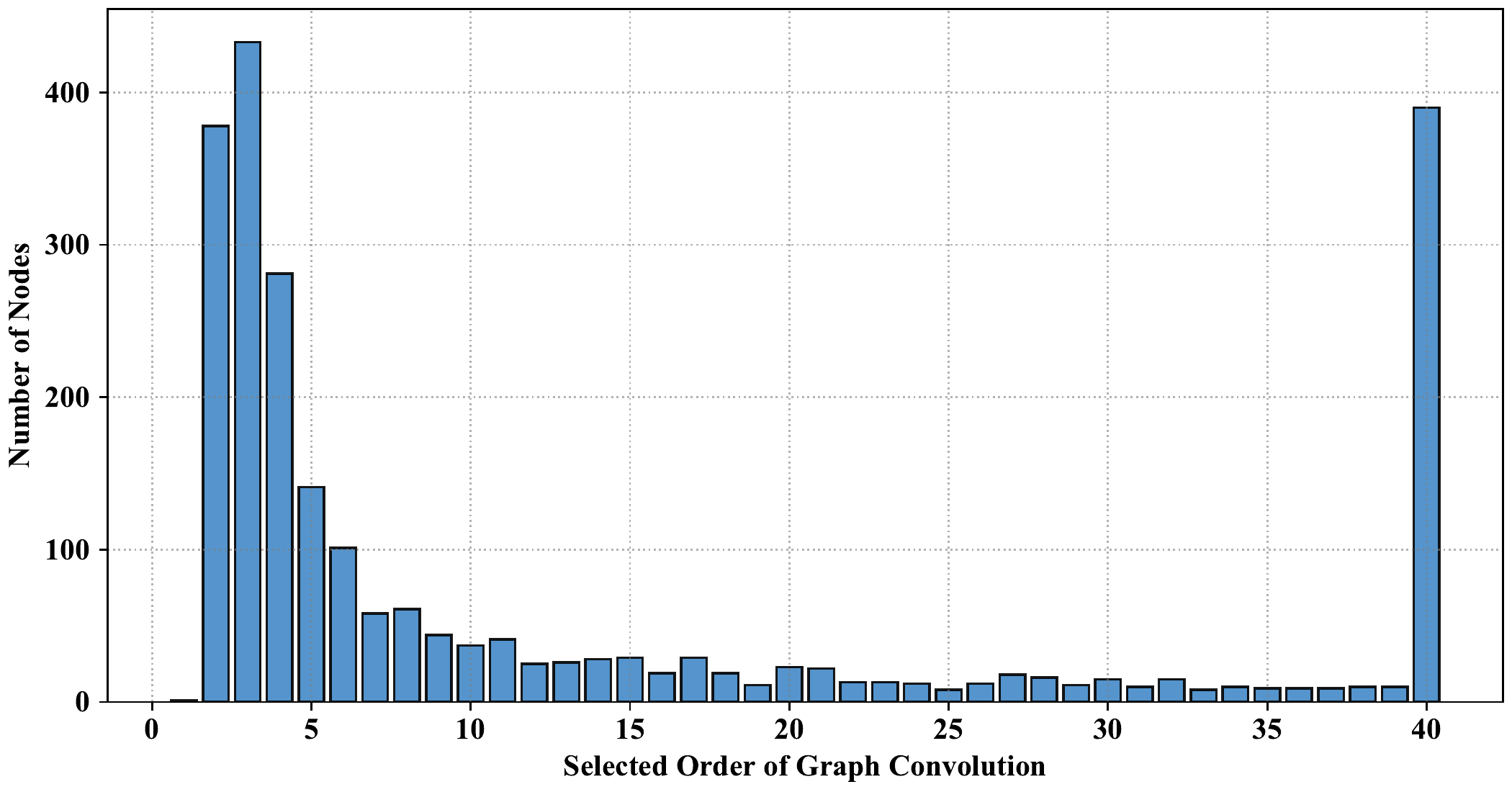}
 \caption{Number of vertices with respect to the selected order of graph convolution for the Wiki dataset.}
 \label{distribution_order_wiki}
\end{figure}

We first trained models on the Cora and Wiki datasets individually and recorded the selected order of graph convolution $N_i$ for each node $v_i$. We then calculated the distribution of the number of corresponding nodes relative to the selected order. The results are depicted in Fig. \ref{distribution_order_cora} and \ref{distribution_order_wiki}.
Please note that our algorithm imposes a maximum value on the order of graph convolution, here set to 40 for both Cora and Wiki.

We can clearly observe that the final selected $N_i$ values differ tremendously among the nodes, and the distribution has a long tail. In Fig. \ref{distribution_order_cora}, on Cora, although saturation can be reached for most nodes with an order of graph convolution that is below approximately 23, some nodes still require deeper convolution, even deeper than the maximum order of 40. These two distributions prove the necessity and effectiveness of our proposed node-wise mechanism.

Furthermore, we can also gain some interesting clues from these distributions regarding the possible reasons for the performance difference between our work and previous study.
In the distribution obtained on Cora, the nodes for which the selected order of graph convolution is below 12 account for 75.44\% of the whole graph. This result is surprisingly consistent with AGC, which chooses 12 as the order of graph convolution. Nevertheless, our method chooses to halt convolution at exactly the 12th order for only 3.43\% of the nodes, while for the others, graph convolution is terminated at an order below 12.
From another perspective, the results reveal that 24.56\% of the nodes still require further convolution. However, if a higher order of graph convolution were to be applied in AGC, this would result in oversmoothness and worsen the performance in the clustering task because AGC determines the order of graph convolution only at the graph level.
From Fig. \ref{distribution_order_wiki}, we can also learn the reason why the MGVAE method can achieve a relatively high performance on Wiki compared with the other baselines. The frequency $N(i)$ peaks at 3, and then a sharp drop occurs, which is consistent with the selected order of graph convolution in the MGVAE method.

\subsubsection{What is the efficiency of our proposed method?}

\begin{figure}
 \centering
 \includegraphics[scale=0.21]{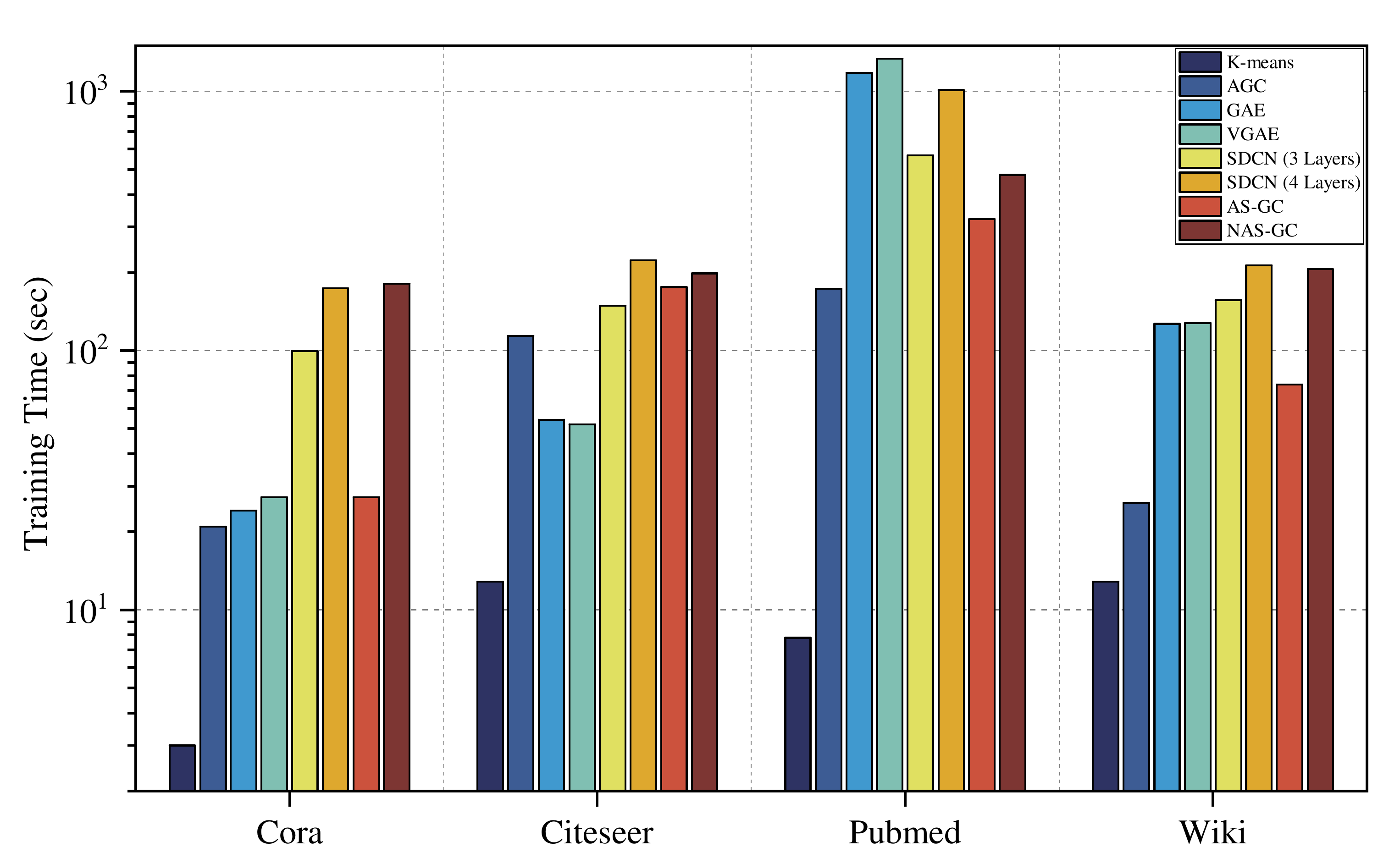}
 \caption{Training times for the baselines and our proposed methods.}
 \label{training_time}
\end{figure}

To compare the time efficiency among the baselines, we present experiments conducted to quantify the training time for each method. Our hardware environment consists of two 3.00 GHz Intel Xeon E5-2687W CPUs and 512 GB of memory. The detailed training times are depicted in Fig. \ref{training_time}. It can be observed that the classical clustering method, k-means, has the lowest time consumption across all datasets and models, followed by AGC, in which no neural network parameters need to be trained. By contrast, there is a common characteristic of other deep learning methods --- many parameters need to be learned. To be concrete, the GAE and VGAE models can be trained most quickly on the Cora and Citeseer datasets, respectively. AS-GC takes the least time on Pubmed and Wiki, followed by NAS-GC on the Pubmed dataset.
Although it seems that the utilization of more fine-grained training targets --- the nodes --- instead of the entire graph in NAS-GC may lead to extra computations, the fact is that these more precise training units can be tremendously beneficial for speeding up the process of convergence.
This superiority is especially obvious for large-scale graph structures, e.g., the Pubmed dataset, which is a graph network composed of 19,717 nodes and 44,338 edges.

\subsubsection{Does the excellent performance of NAS-GC heavily depend on delicate bias hyperparameters?}

The parameters $\lambda_{tig}$ and $\lambda_{sep}$ control the tradeoff between the tightness and separation of the clusters in a dataset. When $\lambda_{tig}$ is equal to 1, a larger $\lambda_{sep}$ places a higher emphasis on a greater separation among clusters, while a smaller $\lambda_{sep}$ focuses on a closer intra-cluster relationship. Therefore, this hyperparameter is responsible for adapting this tradeoff to various graph networks with different characteristics. The problem addressed here is whether the performance of NAS-GC is tightly correlated with the selection of these bias hyperparameters.

\begin{table}[!htbp]
  \renewcommand{\arraystretch}{1.3}
  \caption{Performance of NAS-GC with various $\lambda_{sep}$ values}
  \centering
  \begin{tabular}{c|ccc|ccc}
    \hline
    \multicolumn{1}{c|}{\multirow{2}{*}{Method}}& \multicolumn{3}{c|}{Citeseer} & \multicolumn{3}{c}{Wiki}\\
  \multicolumn{1}{c|}{}&Acc\%&NMI\%&F1\%&Acc\%&NMI\%&F1\%\\
    \hline
    Best & 67 & 41.13 & 62.48 & 50.14 & 47.97 & 40.36 \\
    \hline
    1:3 & \textbf{69.58} & \textbf{43.78} & \textbf{64.73} & 50.89 & 50.13 & 42.95 \\
    1:4 & 68.41 & 42.76 & 63.62 & 51.88 & 50.15 & 42.79 \\
    1:5 & 68.53 & 42.64 & 63.88 & 53.16 & 50.81 & 44.79 \\
    1:10 & 68.86 & 43.16 & 64.06 & \textbf{56.24} & \textbf{51.07} & \textbf{44.79} \\
    1:20 & 68.62 & 42.97 & 63.84 & 55.39 & 51.68 & 44.93 \\
    1:30 & 68.74 & 43.09 & 63.95 & 53.86 & 48.99 & 42.22 \\
    1:40 & 68.68 & 43.07 & 63.89 & 52.07 & 48.74 & 41.68 \\
    1:50 & 68.71 & 43.11 & 63.92 & 53.42 & 49.13 & 42.01 \\
    \hline
    \end{tabular}
  \label{performance_bias_parameter}
\begin{tablenotes}
\item[1] \textbf{note:} ``Best’’ denotes the baseline with the best performance. Bold type indicates the optimal scores.
\end{tablenotes}
\end{table}

As mentioned earlier, instead of directly tuning the hyperparameters through painstaking exploration of the dataset characteristics, we can start by observing the proportion of $\mathcal L_{tig}$ with respect to $\frac{1}{\mathcal L_{sep}}$ after executing the first epoch. We present a group of experiments conducted on Cora and Wiki by adjusting the proportion between the two terms from 1:3 to 1:50 by modifying $\lambda_{tig}$ and $\lambda_{sep}$, with the corresponding performance results recorded in Table \ref{performance_bias_parameter}. A grid search on the proportion sequence can automatically reveal the optimal hyperparameters settings, indicated with bold font in this table. We also noted that within this proportion range, NAS-GC, the proposed method, consistently outperforms the baseline method with the best performance --- either AGC or MGAE. This consistently excellent performance proves that the proposed self-supervision strategy indeed seeks a ``best'' setting, not merely a ``good'' setting, and the excellent performance of NAS-GC is robust.

\subsubsection{Can the learned representations provide an intuitive visualization?}
\begin{figure*}
  \centering
  \subfigure[Before training]{\includegraphics[width=1.1in]{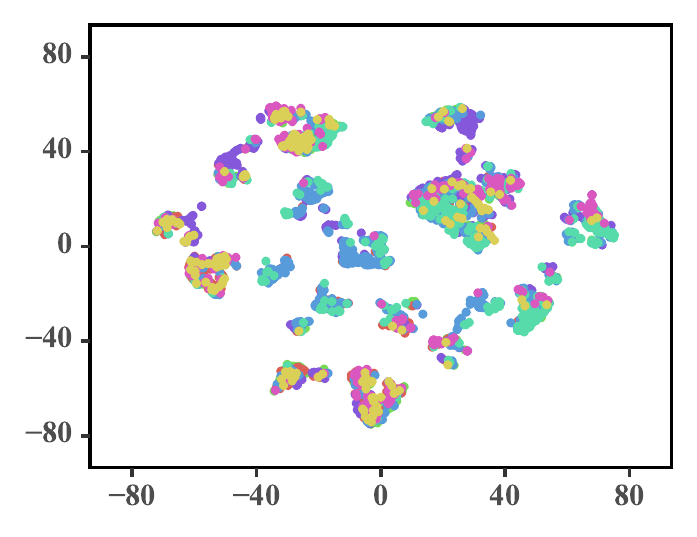}}
  \subfigure[Epoch 1]{\includegraphics[width=1.1in]{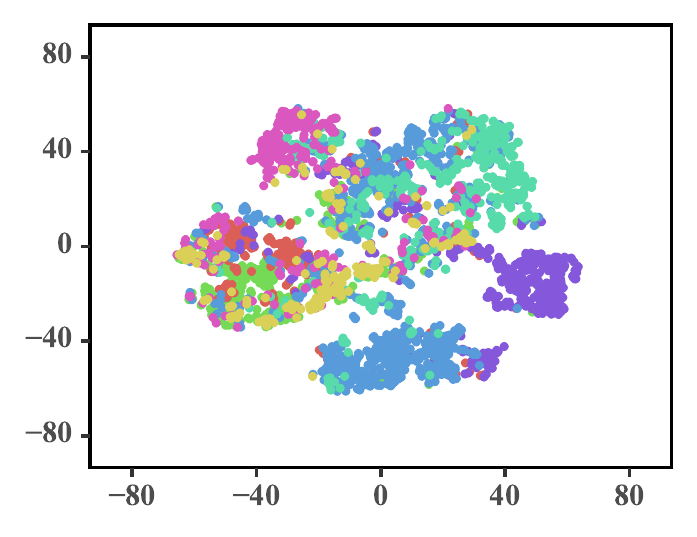}}
  \subfigure[Epoch 20]{\includegraphics[width=1.1in]{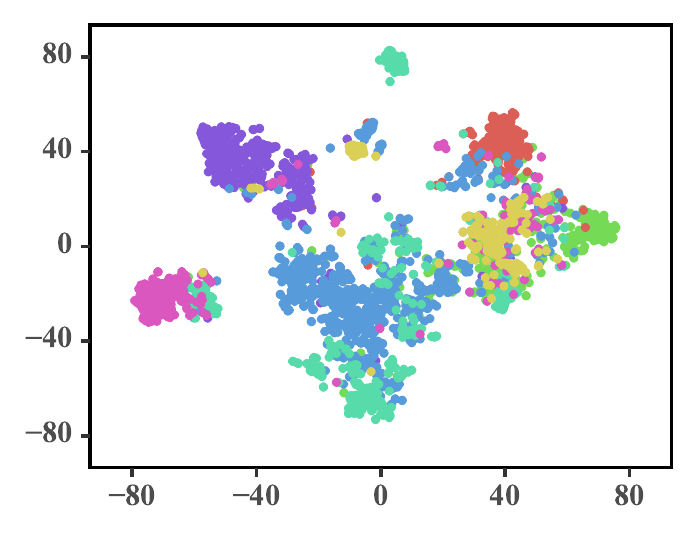}}
  \subfigure[Epoch 50]{\includegraphics[width=1.1in]{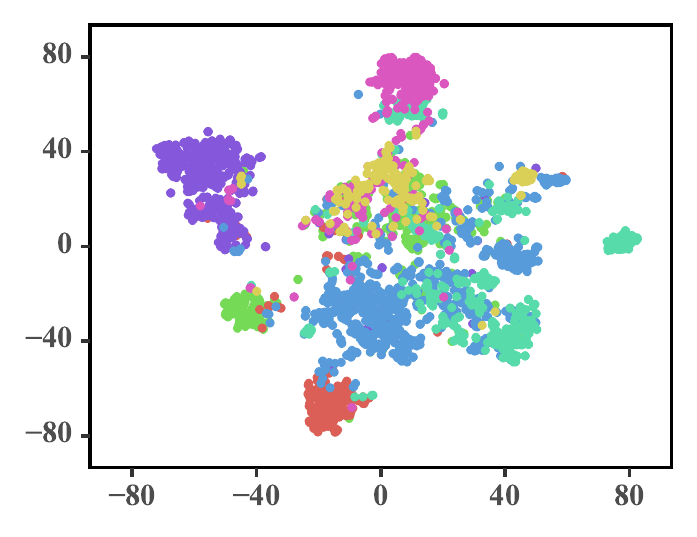}}
  \subfigure[Epoch 100]{\includegraphics[width=1.1in]{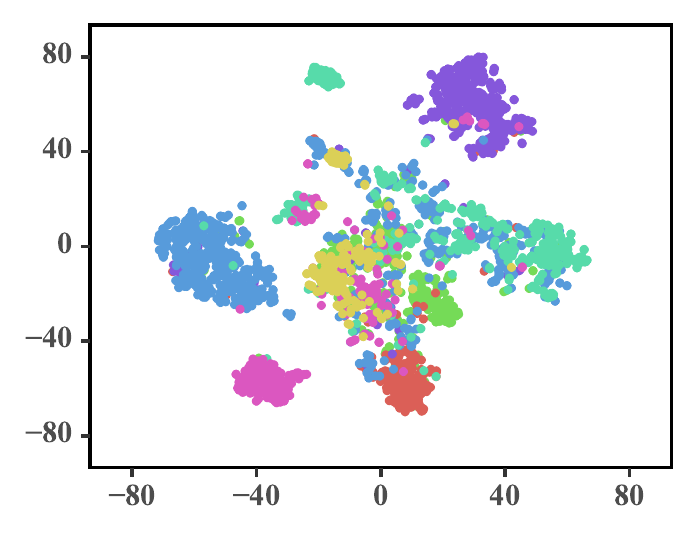}}
  \subfigure[Epoch 200]{\includegraphics[width=1.1in]{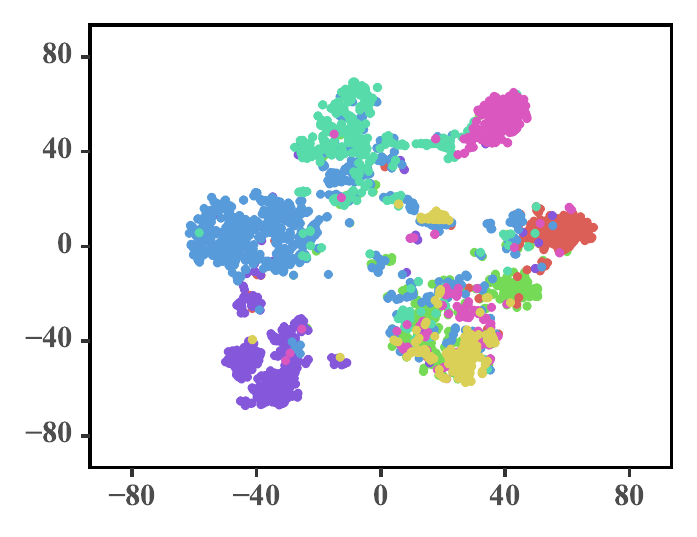}}
  \caption{2D visualizations of the learned representations from NAS-GC with an increasing number of epochs.}
  \label{clustering_visualization}
\end{figure*}

A meaningful visualization of the learned representations can usually provide an indicator for qualifying the learned representations. We select the Cora dataset for such visualization because of its moderate number of classes.

We first map Cora into a low-dimensional space, project the learned representations into a similarity matrix through a linear kernel, and then apply the t-SNE algorithm \cite{maaten2008visualizing} to further project the representations into a 2-dimensional space.
To show the gradual training process, we present visualizations of the representations learned in different epochs.

The results are depicted in Fig. \ref{clustering_visualization}, in which 7 different colors denote the 7 originally labeled document classes. We can observe that the clustering effect already seems reasonable, with separated pink and purple clusters, when the number of epochs reaches 20. However, there are still overlaps among the cyan and blue groups. These overlaps remain until approximately the 200th epoch, when the cyan and blue nodes are approximately separated, although the clusters colored in red, green and yellow still cannot be clearly separated yet. It can be concluded that these three categories are similar enough that they cannot be split in 2-dimensional space. The subjects of genetic algorithms (red), reinforcement learning (green), and theory (yellow) are distinct, yet inevitably connected to each other.

\section{Conclusion and Future Work}
Oversmoothing of GCNs will cause the nodes of a graph to be grouped into fewer clusters and thus poses a challenge in terms of performance degradation. This paper proposes a solution to overcome the oversmoothing of GCNs and the resulting performance degradation of downstream clustering for attributed graphs.
Convolution at a fixed order $k$ at the graph level tends to cause either undersmoothing or oversmoothing.
In this study, we explore how $k$-order filtered graph signals can be evolved via a transition from $(k-1)$-order filtered signals in terms of smoothness.
We design a smoothness sensor to sense the graph smoothness and terminate the graph convolution process once the smoothness is saturated.
Furthermore, we propose a node-wise smoothness-transition mechanism by adaptively customizing the order $k$ of graph convolution for each node.
Finally, a clustering criterion considering both the tightness within clusters and the separation between clusters is defined as the loss function to guide the training of the whole model. Experiments prove that the proposed methods significantly outperform 12 other state-of-the-art baselines in terms of three different metrics across four benchmark datasets. In addition, an extensive study reveals the reasons for their effectiveness and efficiency. In the future, the potential of node-wise smoothness can be further exploited for other downstream tasks of GCNs, such as node classification.

\ifCLASSOPTIONcaptionsoff
  \newpage
\fi

\bibliographystyle{IEEEtran}
\bibliography{references}

\end{document}